\documentclass[10pt,twocolumn,letterpaper]{article}

\usepackage{iccv}              

\usepackage{multirow}
\usepackage{xcolor,colortbl}
\definecolor{myrowcolor}{RGB}{206, 224, 241}

\usepackage[accsupp]{axessibility}
\usepackage{amsmath}
\usepackage{amssymb}
\usepackage{mathtools}
\usepackage{amsthm}
\usepackage{algorithm}
\usepackage{algorithmic}

\definecolor{iccvblue}{rgb}{0.21,0.49,0.74}
\usepackage[pagebackref,breaklinks,colorlinks,allcolors=iccvblue]{hyperref}

\def\paperID{10244} 
\def\confName{ICCV}
\def\confYear{2025}

\title{PLAN: Proactive Low-Rank Allocation for Continual Learning}

\author{
    Xiequn Wang\thanks{Equal contribution.}\textsuperscript{\rm ~~1}~~~
    Zhan Zhuang\footnotemark[1]\textsuperscript{\rm ~~1,2}~~~
    Yu Zhang\thanks{Corresponding author.}\textsuperscript{\rm ~~1} \\
    \textsuperscript{\rm 1} Southern University of Science and Technology \hspace{0.3cm}
    \textsuperscript{\rm 2} City University of Hong Kong \\
     \vspace{1ex} 
    {\tt\small \{wangxiequn,yu.zhang.ust\}@gmail.com, 12250063@mail.sustech.edu.cn}
}
\begin{document}
\maketitle
\begin{abstract}
Continual learning (CL) requires models to continuously adapt to new tasks without forgetting past knowledge. In this work, we propose \underline{P}roactive \underline{L}ow-rank \underline{A}llocatio\underline{N} (PLAN), a framework that extends Low-Rank Adaptation (LoRA) to enable efficient and interference-aware fine-tuning of large pre-trained models in CL settings. PLAN proactively manages the allocation of task-specific subspaces by introducing orthogonal basis vectors for each task and optimizing them through a perturbation-based strategy that minimizes conflicts with previously learned parameters. Furthermore, PLAN incorporates a novel selection mechanism that identifies and assigns basis vectors with minimal sensitivity to interference, reducing the risk of degrading past knowledge while maintaining efficient adaptation to new tasks. Empirical results on standard CL benchmarks demonstrate that PLAN consistently outperforms existing methods, establishing a new state-of-the-art for continual learning with foundation models.
\end{abstract}    
\section{Introduction}
\label{sec:intro}
Continual learning (CL)~\cite{parisi2019continual,van2019three,wang2024comprehensive}, also known as incremental learning or lifelong learning, is a learning paradigm in which a model processes and learns a sequence of tasks while preventing the catastrophic forgetting~\cite{french1999catastrophic,mccloskey1989catastrophic} of previously acquired knowledge. CL plays a critical role in real-world applications such as autonomous driving~\cite{shaheen2022continual,verwimp2023clad,meng2025preserving} and robotics~\cite{she2019openlorisobject,lesort2020continual}, where models must continuously adapt to evolving and non-stationary environments. However, this setting inherently faces the stability–plasticity dilemma~\cite{kirkpatrick2017overcoming, wu2021striking, kim2023stability}, requiring models to maintain stability on previously learned tasks while preserving sufficient plasticity to effectively acquire new information.

With the rise of large-scale pre-trained models, CL has found a strong foundation in freezing transferable representations, which helps preserve general knowledge and mitigate forgetting across tasks. Building on this, parameter-efficient fine-tuning (PEFT) techniques have emerged as promising solutions for CL~\cite{wang2022dualprompt, wang2022learning, smith2023coda, liang2024inflora, wang-etal-2023-orthogonal}. By introducing only a small number of task-specific parameters, PEFT methods such as adapters~\cite{houlsby2019parameter}, prompt tuning~\cite{DBLP:conf/emnlp/LesterAC21}, and low-rank adaptation (LoRA)~\cite{hu2021lora} enable new tasks to be learned efficiently with minimal disruption to existing knowledge. These lightweight modules as task vector~\cite{ilharco2022editing,wistuba2023continual} incrementally encode task-specific updates while preserving the backbone's shared representations, making PEFT a natural fit for continual learning.

\begin{figure}[t]
    \centering
    \vspace{0.15in}
    \includegraphics[width=\linewidth]{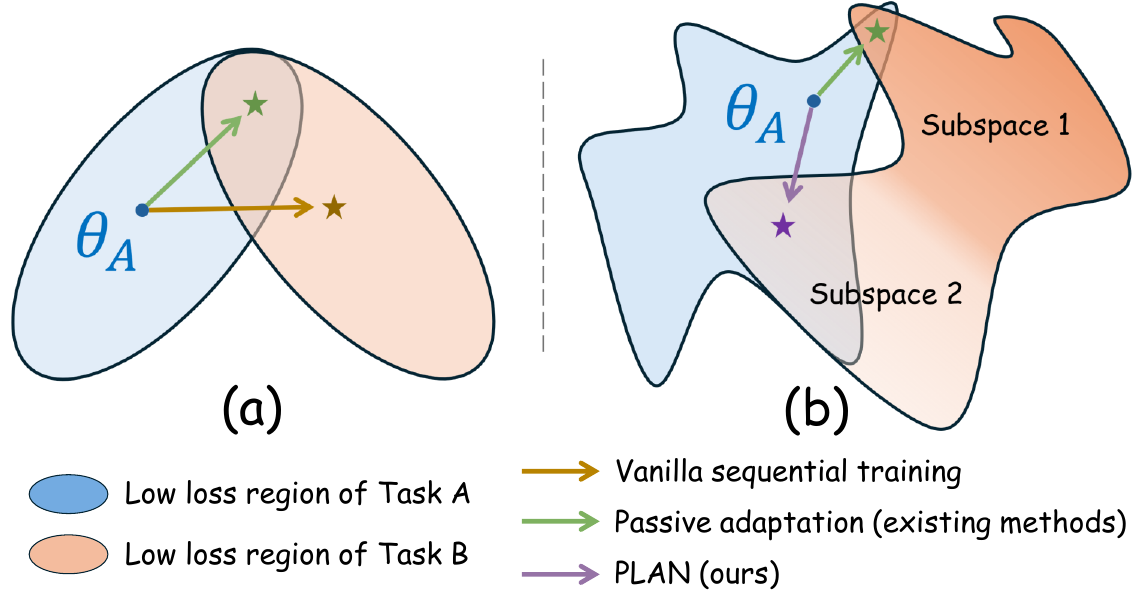}
    \caption{Conceptual illustration of PLAN compared to existing approaches.
    (a) Vanilla sequential training (orange arrow) causes parameter interference, moving the model away from previously low-loss regions.
    Existing methods (green arrow) passively avoid interference by enforcing orthogonality, typically assuming simplified low-loss regions for each task.
    (b) In practice, tasks possess multiple optimal regions. PLAN (purple arrow) is the first method to proactively optimize task-specific subspaces, explicitly anticipating future interference and robustly preserving performance within favorable regions for both current and previous tasks.}
    \label{fig:enter-label}
    \vspace{0.05in}
\end{figure}
Among PEFT methods, LoRA~\cite{hu2021lora} is highly effective, but applying it to CL is non-trivial. A simple strategy of allocating a separate LoRA module per task still faces a critical challenge: \textit{how to ensure that updates for new tasks do not indirectly degrade the performance of previously learned tasks}?

Existing works like O-LoRA~\cite{wang-etal-2023-orthogonal} and InfLoRA~\cite{liang2024inflora} address this by enforcing orthogonality constraints, isolating new task updates from old ones. However, these methods adopt fundamentally \textbf{passive} strategies. Their primary goal is to prevent interference by avoiding shared subspaces, rather than actively identifying subspaces that are inherently more robust to future changes. This passive stance, while effective at reducing forgetting, limits the potential for more strategic adaptation.

To address this limitation, we propose the \underline{P}roactive \underline{L}ow-rank \underline{A}llocatio\underline{N} a novel LoRA-based CL method that shifts from passive isolation to \textbf{proactive subspace planning}. Instead of merely preventing task conflicts, PLAN explicitly anticipates future interference during the training of each task and actively allocates subspaces to minimize potential conflicts across the entire task sequence. PLAN introduces two innovative mechanisms: \textbf{(1)} a perturbation-based optimization objective that anticipates worst-case interference scenarios during current task training, and \textbf{(2)} an orthogonal basis selection strategy informed by this objective, which proactively identifies optimal, low-interference directions in the parameter space for future tasks.

Specifically, each task-specific update in PLAN is represented via a low-rank decomposition ($\Delta W_t = B_t A_t$). PLAN first selects a fixed orthogonal basis for $A_t$ from a predefined set, ensuring new tasks occupy distinct subspaces. Then, PLAN optimizes $B_t$ using a novel min-max objective that robustly prepares the model against worst-case perturbations in the parameter directions reserved for future tasks. This forward-looking process not only makes the current task's parameters more robust but also guides the selection of the most stable basis vectors for the \textit{next} task.

\begin{figure*}[t]
\centering
\includegraphics[width=\linewidth]{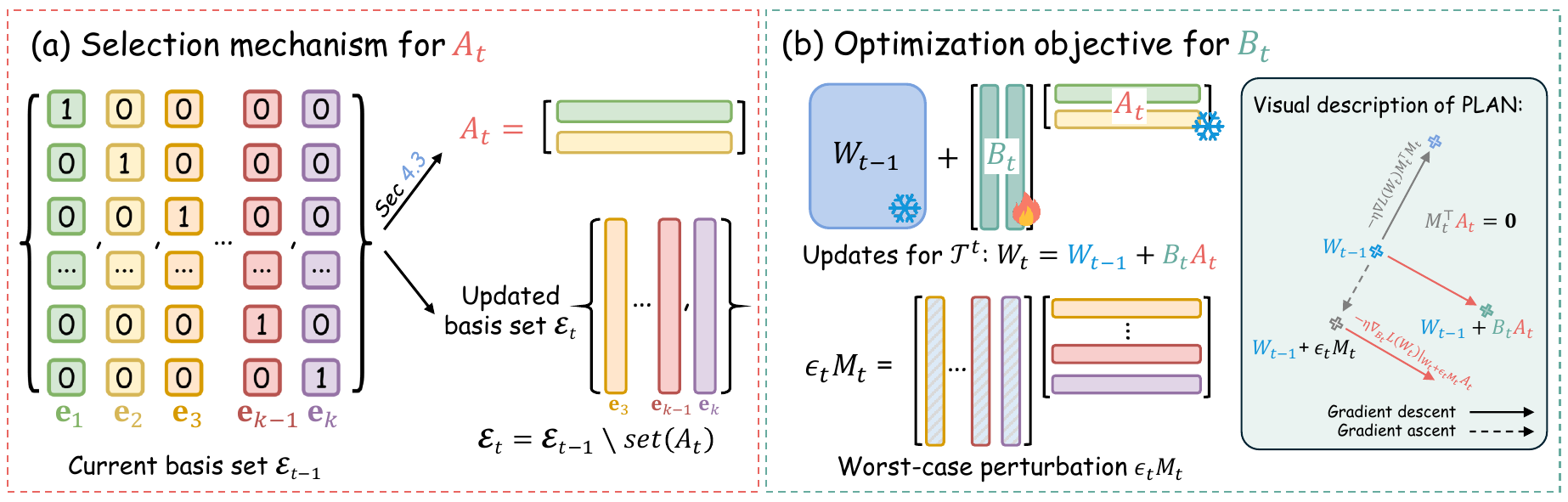}
\caption{Overview of the proposed PLAN method. (a)~Selection mechanism for $A_t$. For clarity, the case $t=1$ is shown, where the current basis set $\mathcal{E}_0$ is the complete standard orthogonal basis. (b)~Optimization objective for $B_t$ using worst-case perturbations along unselected basis vectors, \ie those not included in $A_t$, (denoted by $\mathcal{E}_t = set(M_t)$), with the gradient update directions visually indicated.}
\label{fig:architecture}
\end{figure*}

In summary, PLAN first advances LoRA-based continual learning through proactive interference mitigation, enabling robust lifelong adaptation of pre-trained models. Our contributions are summarized as follows.

\begin{itemize}
\item We shift LoRA-based CL from passive avoidance to proactive planning, where PLAN anticipates future conflicts and strategically selects task-specific subspaces.
\item We introduce a min-max optimization strategy that anticipates worst-case perturbations, ensuring robust parameter updates and improved knowledge retention.
\item We propose an efficient orthogonal basis selection mechanism that eliminates the need for additional subspace learning while maintaining a structured and interference-free representation space across tasks.
\item Through extensive experiments, we show that PLAN surpasses existing CL methods across multiple benchmarks.
\end{itemize}

\section{Related Work}
\label{sec:relatedwork}

\noindent{\bf Continual Learning with PEFT.} 
With the advent of large pre-trained models, recent CL approaches increasingly adopt PEFT to mitigate forgetting.  Among these, prompt-based methods such as L2P~\cite{wang2022learning}, DualPrompt~\cite{wang2022dualprompt}, and CODA-Prompt~\cite{smith2023coda} introduce small trainable prompts for each task for ViTs~\cite{dosovitskiy2020image}. However, these approaches face scalability challenges: as the number of tasks increases, the prompt pool grows linearly, requiring additional mechanisms to select appropriate prompts during inference. Moreover, with long task sequences, prompts tend to become homogeneous~\cite{gao2023unified}. 

In contrast, LoRA-based methods~\cite{wang-etal-2023-orthogonal, liang2024inflora, wistuba2023continual, wei2024online, wu2025sdlora} offer a more scalable alternative by introducing lightweight, task-specific weight updates. For example, O-LoRA~\cite{wang-etal-2023-orthogonal} incrementally learns new tasks in subspaces orthogonal to all previous ones by constraining gradient updates to lie in the null space of past LoRA directions, thereby preventing interference without revisiting prior data. Similarly, InfLoRA~\cite{liang2024inflora} constructs interference-free subspaces by enforcing orthogonality constraints on low-rank adapters, ensuring that new task updates remain nearly orthogonal to those of earlier tasks. Both methods demonstrate the effectiveness of subspace isolation, showing that carefully selecting update directions can significantly reduce interference and achieve a stability–plasticity trade-off. Building on these insights, our work further explores subspace allocation, shifting from passive isolation toward more proactive strategies.

\noindent{\bf Min-Max Optimization in CL.} The min-max objective has been explored to enhance robustness in CL. A notable example is FS-DGPM~\cite{deng2021flattening}, which uses Sharpness-Aware Minimization (SAM)~\cite{foretsharpness} to flatten the loss landscape of \textit{past} tasks, making them more resilient to parameter changes. PLAN's objective is fundamentally different: instead of reacting to preserve past knowledge, it \textbf{proactively} perturbs the parameter space reserved for \textit{future} tasks. This forward-looking optimization finds a solution for the \textit{current} task that is already robust to anticipated future updates, and crucially, it informs the selection of the most stable subspaces for the \textit{next} task.

\noindent{\bf Adaptive LoRA Techniques.} The architecture of LoRA has been extensively refined for better efficiency and flexibility. Some works reduce parameter counts by sharing matrix components~\cite{ruiz2024hyperdreambooth, zhang2023lora, tian2024hydralora}, while others enhance adaptability by decomposing weight updates into magnitude and direction~\cite{liu2024dora, wu2025sdlora}. Initialization strategies have also been explored to better align with model properties or gradients~\cite{meng2024pissa, wang2024lora}. In contrast to these general-purpose improvements, PLAN introduces two innovations specifically for the continual learning challenge: 1) using a standard orthogonal basis for strict, interference-free subspace allocation, and 2) a proactive optimization strategy that enables forward-looking subspace planning.
\section{Preliminaries}
\noindent{\bf Problem Definition.} Continual learning is formulated as the process of sequentially learning a series of tasks $\mathcal{T} = \{ \mathcal{T}^1, \mathcal{T}^2, \dots, \mathcal{T}^N \}$, where each task  $\mathcal{T}^t$  comprises a training dataset  $\mathcal{D}_t = \{ (\boldsymbol{x}_i^t, y_i^t) \}_{i=1}^{N_t}$  with $N_t$ inputs $\{\boldsymbol{x}_i^t\}$  and the corresponding labels $\{y_i^t\}$. The primary objective of CL is to learn each task in sequence without incurring catastrophic forgetting of previously acquired knowledge.

Formally, for a model parameterized by $\theta$, the goal of continual learning is to minimize the average generalization error over all encountered tasks:
\begin{equation}
\mathcal{L}_{CL}(\theta) = \frac{1}{j} \sum_{i=1}^{j} \mathcal{L}_i(\theta),
\end{equation}
where $j$ denotes the index of the current task, and $\mathcal{L}_i(\theta)$ represents the generalization error of task $\mathcal{T}^i$. We use $L_S(\theta)$ to denote the empirical loss of the model on a dataset $S$.
The model aims to perform well on both the current task $\mathcal{T}_j$ and all previously learned tasks. This paper focuses on class-incremental learning scenarios, using a pre-trained Vision Transformer (ViT)~\cite{dosovitskiy2020image} as the initial classification model.

\vspace{0.05in}
\noindent{\bf Low-Rank Adaptation.} LoRA is a parameter-efficient fine-tuning technique for large pre-trained models. Given a pre-trained weight matrix $W_0 \in \mathbb{R}^{d \times k}$ from the model, LoRA introduces two low-rank matrices $B \in \mathbb{R}^{d \times r}$ and $A \in \mathbb{R}^{r \times k}$, where the rank $r$ is chosen to be much smaller than $d$ and $k$. The weight update is then formulated as
\begin{equation}
W_{\text{new}} = W_0 + BA,
\end{equation}
with $W_0$ remaining fixed and only $A$  and $B$ being trainable. 




\section{Methodology}
In this section, we introduce the proposed PLAN method.

\subsection{Overview}
Building upon LoRA, we train a dedicated LoRA adapter for each task while keeping the pre-trained weights and all previously learned adapters fixed. For simplicity, we use a single MLP layer for illustration. Let $W_0$ denote the pre-trained weight and $W_{t-1}$ the updated weight before learning from task $\mathcal{T}^{t}$. The weight update with the newly added LoRA for task $\mathcal{T}^{t}$ is then formulated as
\begin{equation}
W_t = \underbrace{W_0 + \sum_{i=1}^{t-1} B_i A_i}_{\text{frozen}} + B_t A_t = W_{t-1} + B_t A_t,
\end{equation}
where $B_t \in \mathbb{R}^{d \times r_t}$ and $A_t \in \mathbb{R}^{r_t \times k}$, with $r_t$ being the LoRA rank allocated for the new task. As illustrated in Figure~\ref{fig:architecture}, the proposed PLAN method incorporates two key designs to mitigate interference between previously learned and newly acquired knowledge: \textbf{(1)} constructing $A_t$ by selecting a set of orthogonal basis vectors without additional training,
and \textbf{(2)} optimizing $B_t$ by applying perturbations along the directions of the unselected basis vectors.

\vspace{0.05in}
\noindent{\bf Initialization.} We begin by defining a standard basis set $\mathcal{E}_0 = \{\mathbf{e}_i\}^{k}_{i=1}$, where each $k$-dimensional vector $\mathbf{e}_i$ has all entries set to zero except for the $i$-th entry, which is set to $1$. For each task $\mathcal{T}^{t}$, we select $r_t$ basis vectors to form $A_t$ from the current basis set and update the basis set as 
\begin{equation}
\mathcal{E}_t = \mathcal{E}_{t-1} \setminus set(A_t),
\end{equation}
where $set(A_t)$ denotes the set of row vectors of $A_t$ selected from $\mathcal{E}_{t-1}$, and the operation $\setminus$ indicates the removal of vectors from a set. For the first task, we construct $A_1$ by selecting the first $r_1$ basis vectors, \ie, $A_1 = \mathbf{e}_{[1:r_1]}^\top$. We detail the selection strategy for $A_t$ with rank $r_t$ for subsequent tasks in Section~\ref{sec:method-selectA} and describe the optimization of $B_t$ for improved future allocation is described in Section~\ref{sec:method-optimizeB}.

In practice, the total number of basis vectors $k$ corresponds to the feature dimension of the model, which is typically large and significantly greater than the number of tasks encountered in continual learning scenarios. As a result, the basis set $\mathcal{E}_t$ rarely becomes empty in realistic settings, ensuring sufficient subspace capacity for long task sequences.

\subsection{Optimization Objective for $B_t$}\label{sec:method-optimizeB}
For task $\mathcal{T}^{t}$, once $A_t$ is determined to allocate a task-specific subspace, we optimize the corresponding parameters in $B_t$
to effectively capture task-specific information. 

In this section, we introduce a min-max optimization objective that offers two key advantages. First, it proactively mitigates parameter conflicts across tasks by optimizing the current task's matrix $B_t$ to be robust against interference from future tasks. Second, by explicitly analyzing perturbation sensitivity along different basis-vector directions, our method provides critical insights for selecting optimal initial basis vectors for subsequent tasks, a strategy we describe in detail in Section~\ref{sec:method-selectA}. 


In the proposed PLAN method, basis vectors are orthogonal, meaning that $\mathbf{e}_i^\top \mathbf{e}_j = 0$ for $i \neq j$. Since each $A_t$ is constructed from a unique set of these basis vectors, $A_t$ is of full rank and remains orthogonal to $A_i$ from other tasks (\ie, $A_i^\top A_j = 0$ for $i \neq j$). Consequently, the set $\{A_t\}_{t=1}^M$ projects the input onto distinct subspaces, allowing the model to learn new task features without interfering with previously acquired knowledge.

Let $M_t$ denote the matrix formed by the row vectors corresponding to the unselected basis set $\mathcal{E}_t$:
\begin{equation}
M_t = \text{span}(\mathcal{E}_t).
\end{equation}
Since a future task $\mathcal{T}^{s}$ $(s>t)$ will select its $A_s$ from the remaining rows of $M_t$, we introduce a min-max optimization for $B_t$ to proactively reduce potential interference from future tasks. The objective is formulated as:
\begin{equation}
\min_{B_t} \max_{\|\epsilon_t \|_p \leq \rho} L_{\mathcal{D}_t}(W_{t-1}+ B_tA_t +  \epsilon_t M_t),\label{obj_fun}
\end{equation}
where $\rho$ is a hyperparameter controlling the perturbation magnitude, $|M_t|$ is the number of basis vectors in $M_t$, $\epsilon_t \in \mathbb{R}^{d \times |M_t|}$ represents the worst-case perturbation applied to the unallocated subspace $M_t$, and $\|\cdot\|_p$ denotes the $\ell_p$ norm of a vector after transforming matrices to vectors. Here the term $\epsilon_tM_t$ models represents the possible interference by future tasks.

To solve problem \eqref{obj_fun}, we approximate the inner maximization problem via the first-order Taylor expansion as
\begin{align}\label{eq:bhat}
&\arg\max_{\|\epsilon\|_p \leq \rho}
L_{\mathcal{D}_t}\big(W_t + \epsilon M_t\big) \notag \\
\approx &\arg\max_{\|\epsilon\|_p \leq \rho}
\left[L_{\mathcal{D}_t}(W_t) \notag + \epsilon^\top \nabla_{W_t} L_{\mathcal{D}_t}(W_t) M_t^\top \right]\notag \\
= &\arg\max_{\|\epsilon\|_p \leq \rho}
\epsilon^\top \nabla_{W_t} L_{\mathcal{D}_t}(W_t) \, M_t^\top. \end{align}
Problem \eqref{eq:bhat} has a closed-form solution as
\begin{equation}\label{eq:solveepsilon}
\hat{\epsilon}_t(W_t) = \rho \,\frac{|\mathbf{g}|^{\,q-1} \odot \mathrm{sign}(\mathbf{g})}{(\|\mathbf{g}\|_q^{\,q})^{\frac{1}{p}}},
\end{equation} 
where $\mathbf{g}= \nabla_{W_t} L_{\mathcal{D}_t}(W_t)M_t^\top$, $q$ satisfies $\frac{1}{p} + \frac{1}{q} = 1$, $|\mathbf{g}|$ denotes the elementwise absolute value of $\mathbf{g}$, the operator $\odot$ indicates elementwise multiplication, and $\mathrm{sign}(\mathbf{g})$ returns the elementwise sign of $\mathbf{g}$.
By plugging $\hat{\epsilon}_t(W_t)$ into problem \eqref{obj_fun}, 
the objective for ${B_t}$ is formulated as
\begin{equation}
\min_{B_t} L_{\mathcal{D}_t}(W_{t-1}+ B_tA_t +  \hat{\epsilon}_t(W_t)  M_t).
\label{obj_B}
\end{equation}
To avoid the computation of the Hessian matrix and reduce the computational cost, we treat $\hat{\epsilon}_t(W_t)$ as a constant instead of a function of $W_t$ or $B_t$ and write it as $\hat{\epsilon}_t$. Then we can solve problem \eqref{obj_B} via stochastic gradient descent or its variants with the gradient computed as 
\begin{equation}
\begin{aligned}
&\ \ \  \nabla_{B_t} L_{\mathcal{D}_t}(W_{t-1}+ B_tA_t +  \hat{\epsilon}_t(W_t)  M_t) \\
&\approx \nabla_{B_t} L_{\mathcal{D}_t}(W_t) \big|_{W_{t-1} + B_t A_t + \hat{\epsilon}_tM_t} A_t. 
\end{aligned}
\end{equation}

\subsection{Selection Mechanism for $A_{t+1}$}\label{sec:method-selectA}

After learning $B_t$ as introduced in the previous section,  we now introduce the selection mechanism for $A_{t+1}$ in the proposed PLAN method.
During the training of the previous task $\mathcal{T}^{t}$, we compute perturbations for each mini-batch using the formulation in Eq.~\eqref{eq:solveepsilon}. Specifically, to solve the inner maximization problem at each step $s$, we calculate the 2-norm of each column in the perturbation matrix 
$\hat{\epsilon}_t$ as:
\begin{align}
    n^s_{t, j} = \|\hat{\epsilon}_{t, j}\|_2,
\end{align}
where $\hat{\epsilon}_{t, j}$ denotes the $j$-th column of $\hat{\epsilon}_{t}$.

We then define a frequency function $h(i)$ that counts the number of times index $i$ appears among these $r$ smallest values over the last $S$ training steps (i.e., a sliding window of size $S$):
\begin{equation}\label{eq:select1}
h(i) = \sum_{s'=s-S+1}^{S} \mathbb{I}\Big(i \in \arg\min_{r}\  n^{s'}_{t, j}\  \Big|\ j = 1, \dots, |M_t| \Big), 
\end{equation}
where $\mathbb{I}(\cdot)$ is the indicator function, and $\arg\min_{r}$ returns the set of $r$ indices with the smallest values. The parameter $S$ represents the size of the sliding window for accumulating frequencies. As shown in our analysis in Appendix~\ref{sec:appendix_analysis_s}, a small value for $S$ is sufficient, and we use $S=50$ in our experiments.


\begin{algorithm}[t]
   \caption{PLAN Method for Continual Learning}
   \label{alg:PLAN}
\begin{algorithmic}
   \STATE {\bfseries Input:} a pre-trained ViT model $f_\theta$, number of tasks $T$, training set ${\{\{x_i^t,y_i^t\}_{i=1}^{n_t}\}}_{t=1}^T$, number of training epochs $E$, predefined LoRA basis set $\mathcal{E}_0$. 
   \STATE {\bfseries Output:} The learned LoRA parameters ${\{A_t,B_t\}}_{t=1}^T$.
   \FOR{$t$ in 1, ..., $T$}
   \STATE Construct $A_t$ through  Eqs.~\eqref{eq:select1} and \eqref{eq:select2};
   \STATE $\mathcal{E}_t \leftarrow \mathcal{E}_{t-1} \setminus set(A_t)$;
   \FOR{$e$ = 1, ..., $E$}
   \STATE Sample batch $\mathcal{B} = \{(\boldsymbol{x}_1^t, {y}_1^t), ... (\boldsymbol{x}_b^t, {y}_b^t)\}$;
   \STATE $\mathbf{g} \leftarrow \nabla_{{W_t}} L_\mathcal{B}({W_t})M_t^\top$;
   \STATE Compute $\hat{{\epsilon}}_t({W_t})$ with $\mathbf{g}$ according to Eq.~\eqref{eq:bhat};
   \
   \STATE $\mathbf{g}^{\text{PLAN}} \leftarrow \nabla_{B_t} L_\mathcal{B}({W_t})|_{W_{t-1}+A_t{B_t}+\hat{{\epsilon}}_t(W_t)M_t}$;
   \STATE Update $B_t$ with $\mathbf{g}^{\text{PLAN}}$ through gradient descent;
   \ENDFOR
   \ENDFOR
\end{algorithmic}
\end{algorithm}

These indices with the highest frequencies are considered most significant, as they consistently exhibit minimal perturbation. Consequently, for the subsequent task $\mathcal{T}^{t+1}$, we select the index set $I_{t+1}$ containing the $r$ indices with the highest frequency values to form $A_{t+1}$ as
\begin{equation}\label{eq:select2}
I_{t+1} = \underset{I \subseteq \{1, \dots, |M_t|\},\, |I| = r_{t+1}}{\operatorname*{arg\,max}}\, \sum_{i \in I}\ h(i).
\end{equation}

\begin{table*}[h]
\centering
\resizebox{0.95\linewidth}{!}{
\begin{tabular}{l|cc|cc|cc}
\toprule
\multirow{2}{*}{\textbf{Method}} & \multicolumn{2}{c|}{\textit{ImageNet-R ($N = 5$)}} & \multicolumn{2}{c|}{\textit{ImageNet-R ($N = 10$)}} & \multicolumn{2}{c}{\textit{ImageNet-R ($N = 20$)}} \\
\cline{2-7}
& \raisebox{-0.25ex}[0pt][0pt]{Acc $\uparrow$} & \raisebox{-0.25ex}[0pt][0pt]{AAA $\uparrow$} & \raisebox{-0.25ex}[0pt][0pt]{Acc $\uparrow$} & \raisebox{-0.25ex}[0pt][0pt]{AAA $\uparrow$} & \raisebox{-0.25ex}[0pt][0pt]{Acc $\uparrow$} & \raisebox{-0.25ex}[0pt][0pt]{AAA $\uparrow$} \\
\midrule
L2P & 64.20 \footnotesize{(±0.30)} & 69.25 \footnotesize{(±0.63)} & 62.52 \footnotesize{(±0.41)} & 68.69 \footnotesize{(±0.35)} & 58.63 \footnotesize{(±0.52)} & 65.67 \footnotesize{(±0.33)} \\
Dual-Prompt & 67.43 \footnotesize{(±1.13)} & 71.40 \footnotesize{(±0.85)} & 64.59 \footnotesize{(±1.24)} & 69.59 \footnotesize{(±0.72)} & 60.89 \footnotesize{(±0.62)} & 66.20 \footnotesize{(±0.51)} \\
CODA-Prompt & 74.52 \footnotesize{(±4.25)} & 78.21 \footnotesize{(±2.73)} & 71.58 \footnotesize{(±0.26)} & 76.47 \footnotesize{(±0.28)} & 67.10 \footnotesize{(±0.46)} & 72.38 \footnotesize{(±0.42)} \\
\textit{Inc-LoRA} & 72.36 \footnotesize{(±0.57)} & 79.60 \footnotesize{(±0.27)} & 63.69 \footnotesize{(±0.84)} & 74.54 \footnotesize{(±0.35)} & 52.12 \footnotesize{(±0.72)} & 67.73 \footnotesize{(±0.45)} \\
O-LoRA & 73.12 \footnotesize{(±6.09)} & 77.33 \footnotesize{(±3.67)} & 65.74 \footnotesize{(±0.81)} & 72.89 \footnotesize{(±0.87)} & 59.94 \footnotesize{(±0.82)} & 68.92 \footnotesize{(±0.69)} \\
InfLoRA & 77.09 \footnotesize{(±0.33)} & \textbf{81.96} \footnotesize{(±0.28)} & 74.37 \footnotesize{(±0.54)} & 80.37 \footnotesize{(±0.62)} & 69.83 \footnotesize{(±0.65)} & 76.83 \footnotesize{(±0.54)} \\
\rowcolor{myrowcolor} PLAN (ours) & \textbf{77.79} \footnotesize{(±0.24)} & {81.93} \footnotesize{(±0.63)} & \textbf{75.25} \footnotesize{(±0.42)} & \textbf{80.41} \footnotesize{(±0.56)} & \textbf{71.06} \footnotesize{(±0.42)} & \textbf{77.93} \footnotesize{(±0.56)} \\
\bottomrule
\end{tabular}
}
\vskip -0.05in
\caption{Comparison of different methods on \textit{ImageNet-R} with varying $N$.}
\label{tab:imagenet-r-comparison}
\end{table*}
\begin{figure*}[t]
\centering
\includegraphics[width=1.\textwidth]{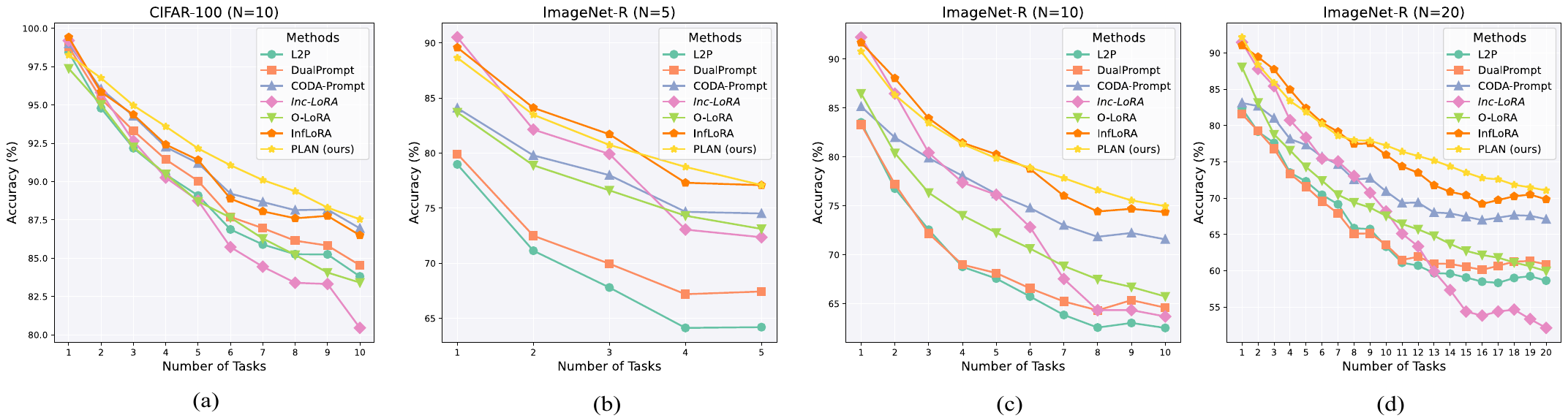}
\vskip -0.05in
\caption{Variation of the performance of different methods during the learning of \textit{ImageNet-R} and \textit{CIFAR100}.}
\label{fig:change-imagenet-r}
\end{figure*}
Finally, $A_{t+1}$ is constructed by selecting the rows of $M_t$ corresponding to the indices in $I_{t+1}$. This selection mechanism ensures that $A_{t+1}$ comprises the basis vectors that consistently experience the least perturbation during training, thereby minimizing interference with previously acquired task knowledge.
We summarize the whole process of our proposed PLAN method in Algorithm~\ref{alg:PLAN}.



\section{Experiments}
This section presents the experimental setup and a comparison of the proposed PLAN method with other continual learning (CL) techniques across multiple benchmarks and foundation models.

\subsection{Experimental Setup}\label{sec:setup}

\noindent \textbf{Datasets.} Following the approach in \cite{liang2024inflora}, we evaluate PLAN using three widely recognized CL benchmarks in the vision domain: CIFAR-100 \cite{krizhevsky2009learning}, DomainNet \cite{peng2019moment}, and ImageNet-R \cite{hendrycks2021many}. CIFAR-100 consists of 100 classes, ImageNet-R includes 200 ImageNet classes rendered in various artistic styles, and DomainNet features 345 classes spanning six distinct domains. For experimental purposes, we divide CIFAR-100 into 10-class subsets, ImageNet-R into tasks containing 40, 20, or 10 classes each (corresponding to 5, 10, or 20 tasks, respectively), and DomainNet into five tasks, each with 69 classes.

\paragraph{Evaluation Protocol.} 
To evaluate the performance of continual learning, we employ two widely used metrics: \textbf{Average Accuracy (Acc)} and \textbf{Average Anytime Accuracy (AAA)}. Acc represents the mean classification accuracy across all tasks at the end of training, providing a final measure of overall performance. In contrast, AAA tracks the cumulative average accuracy over all previously encountered tasks after training on each successive task, offering a more dynamic view of how well the model maintains knowledge throughout the learning process.

\paragraph{Baselines.} The performance of PLAN is compared against several state-of-the-art CL methods, including L2P \cite{wang2022learning}, DualPrompt \cite{wang2022dualprompt}, CODA-Prompt \cite{smith2023coda}, O-LoRA \cite{wang-etal-2023-orthogonal}, and InfLoRA \cite{liang2024inflora}. Additionally, we introduce Incremental LoRA (\textit{Inc-LoRA}) as a baseline to establish a lower bound for LoRA-based methods. \textit{Inc-LoRA} involves training a separate LoRA for each new task and merging it into the original model after each task.

\paragraph{Architectural and Training Details.} 

Following prior studies~\cite{liang2024inflora}, we adopt the ViT-B/16 model, pre-trained on ImageNet-21K and fine-tuned on ImageNet-1K, as our backbone. Additionally, we evaluate our method on the self-supervised ViT-B/16 variant, iBOT~\cite{zhou2021image}, to assess its effectiveness across different training paradigms.

All models are trained using the Adam optimizer~\cite{DBLP:journals/corr/KingmaB14}, which combines running averages of the gradient and squared gradient, with $\beta_1 = 0.9$ and $\beta_2 = 0.999$. Training is conducted for 50 epochs on ImageNet-R, 20 epochs on CIFAR-100, and 5 epochs on DomainNet, with a batch size of 128. In line with \cite{gao2023unified}, LoRA-based methods such as \textit{Inc-LoRA}, O-LoRA \cite{wang-etal-2023-orthogonal}, InfLoRA \cite{liang2024inflora}, and PLAN integrate LoRA modules into the key and value components of the attention mechanism. PLAN introduces a single hyperparameter, $\rho$, which is set to 0.01 across all datasets.

\begin{table*}[h]
\centering
\setlength{\tabcolsep}{12pt}
\resizebox{0.9\linewidth}{!}{
\begin{tabular}{l|cc|cc}
\toprule
\multirow{2}{*}{\textbf{Method}} & \multicolumn{2}{c|}{\textit{CIFAR-100}} & \multicolumn{2}{c}{\textit{DomainNet}} \\
\cline{2-5}
& \raisebox{-0.25ex}[0pt][0pt]{Acc $\uparrow$} & \raisebox{-0.25ex}[0pt][0pt]{AAA $\uparrow$} & \raisebox{-0.25ex}[0pt][0pt]{Acc $\uparrow$} & \raisebox{-0.25ex}[0pt][0pt]{AAA $\uparrow$} \\
\midrule
L2P & 83.81 \footnotesize{(±0.42)} & 89.20 \footnotesize{(±0.36)} & 70.26 \footnotesize{(±0.25)} & 75.83 \footnotesize{(±0.98)} \\
Dual-Prompt & 84.54 \footnotesize{(±0.31)} & 90.02 \footnotesize{(±0.22)} & 68.26 \footnotesize{(±0.09)} & 73.84 \footnotesize{(±0.45)} \\
CODA-Prompt & 86.95 \footnotesize{(±0.36)} & 91.39 \footnotesize{(±0.25)} & 70.58 \footnotesize{(±0.53)} & 76.68 \footnotesize{(±0.44)} \\
\textit{Inc-LoRA} & 80.45 \footnotesize{(±1.20)} & 88.40 \footnotesize{(±0.48)} & 68.26 \footnotesize{(±0.09)} & 73.84 \footnotesize{(±0.45)} \\
O-LoRA & 83.41 \footnotesize{(±0.46)} & 89.05 \footnotesize{(±0.56)} & 70.58 \footnotesize{(±0.53)} & 76.68 \footnotesize{(±0.44)} \\
InfLoRA & 86.50 \footnotesize{(±0.71)} & 91.23 \footnotesize{(±0.38)} & 71.59 \footnotesize{(±0.23)} & \textbf{78.29} \footnotesize{(±0.50)} \\
\rowcolor{myrowcolor} PLAN (ours) & \textbf{87.54} \footnotesize{(±0.31)} & \textbf{92.21} \footnotesize{(±0.35)} & \textbf{72.12} \footnotesize{(±0.16)} & 77.52 \footnotesize{(±0.37)} \\
\bottomrule
\end{tabular}
}
\caption{Comparison of different methods on \textit{CIFAR-100} and \textit{DomainNet} ($N=5$).}
\label{tab:cifar_domainnet}
\end{table*}

\begin{table*}[h]
\centering
\resizebox{0.9\linewidth}{!}{
\begin{tabular}{l|cc|cc|cc}
\toprule
\multirow{2}{*}{\textbf{Method}} & \multicolumn{2}{c|}{\textit{ImageNet-R ($N = 5$)}} & \multicolumn{2}{c|}{\textit{ImageNet-R ($N = 10$)}} & \multicolumn{2}{c}{\textit{ImageNet-R ($N = 20$)}} \\
\cline{2-7}
& \raisebox{-0.25ex}[0pt][0pt]{Acc $\uparrow$} & \raisebox{-0.25ex}[0pt][0pt]{AAA $\uparrow$} & \raisebox{-0.25ex}[0pt][0pt]{Acc $\uparrow$} & \raisebox{-0.25ex}[0pt][0pt]{AAA $\uparrow$} & \raisebox{-0.25ex}[0pt][0pt]{Acc $\uparrow$} & \raisebox{-0.25ex}[0pt][0pt]{AAA $\uparrow$} \\
\midrule
\textit{Inc-LoRA} & 72.36 \footnotesize{(±0.57)} & 79.60 \footnotesize{(±0.27)} & 63.69 \footnotesize{(±0.84)} & 74.54 \footnotesize{(±0.35)} & 52.12 \footnotesize{(±0.72)} & 67.73 \footnotesize{(±0.45)} \\
\midrule
w/o $A_t$ selection & 75.97 \footnotesize{(±0.77)} & 80.69 \footnotesize{(±0.49)} & 72.14 \footnotesize{(±0.51)} & 78.56 \footnotesize{(±0.56)} &68.35 \footnotesize{(±0.53)} & 76.33 \footnotesize{(±1.12)} \\
w/o perturbation & 76.66 \footnotesize{(±0.76)} & 80.38 \footnotesize{(±0.51)} & 74.97 \footnotesize{(±0.40)} & 79.57 \footnotesize{(±0.30)} & 70.65 \footnotesize{(±0.68)} & 76.42 \footnotesize{(±0.76)} \\
\rowcolor{myrowcolor} PLAN (ours) & \textbf{77.79} \footnotesize{(±0.24)} & {\textbf{81.93}} \footnotesize{(±0.63)} & \textbf{75.25} \footnotesize{(±0.42)} & \textbf{80.41} \footnotesize{(±0.56)} & \textbf{71.06} \footnotesize{(±0.42)} & \textbf{77.93} \footnotesize{(±0.56)} \\
\bottomrule
\end{tabular}
}
\caption{Ablation study of our method for two component.}
  \label{table:ablation study}
\end{table*}

\subsection{Main Results}
Table~\ref{tab:imagenet-r-comparison} presents the performance of various methods on ImageNet-R across different task settings, while Table~\ref{tab:cifar_domainnet} shows the results on CIFAR-100 and DomainNet. Our proposed PLAN consistently outperforms previous methods. This improvement can be attributed to PLAN’s proactive and orthogonality-based subspace allocation strategy, which effectively mitigates task interference by proactively selecting distinct parameter subspaces for each task.

In Figure~\ref{fig:change-imagenet-r}, we further illustrate the evolution of accuracy across sequential tasks on ImageNet-R and CIFAR-100. Unlike previous approaches, which exhibit pronounced performance fluctuations and sharp accuracy drops when encountering new tasks, PLAN maintains more stable and higher accuracy levels throughout the learning process.

These empirical results demonstrate that proactively managing parameter update subspaces—rather than passively responding to interference after it occurs—enhances the stability and robustness of continual learning models.

\subsection{Ablation Study}

\paragraph{Ablation Study of PLAN Components.} To validate the contributions of the two main components of PLAN, we conduct experiments to evaluate the effectiveness of our $A_t$ selection mechanism and the optimization strategy for training $B_t$. In the first variant, we modify the $A_t$ selection process to randomly select from $\mathcal{E}_t$. In the second variant, we retain the $A_t$ selection mechanism but remove the $B_t$ optimization algorithm, which is implemented by using the original PLAN method to construct $\mathcal{E}_t$, as the $A_t$ selection mechanism depends on the $B_t$ optimization algorithm. Finally, we remove both components, making the method equivalent to \textit{Inc-LoRA}. Table~\ref{table:ablation study} presents the results of this ablation study, which demonstrate that both components contribute to the effectiveness of PLAN. Both variants outperform \textit{Inc-LoRA}, highlighting the importance of our basis construction for continual learning.

\vskip 0.09in
\noindent{\bf Variants of Basis Set Initialization.}
We evaluate three variants for constructing the basis set $\mathcal{E}$: random orthogonal basis, LoRA-GA initialization~\cite{wang2024lora} and the standard orthogonal basis (ours).

For the random orthogonal basis, $ \mathcal{E} $ is generated by sampling a matrix from a standard Gaussian distribution and then orthogonalizing it using the Gram-Schmidt process~\cite{strang2006linear}. In LoRA-GA initialization, we first compute $ \nabla_{W_0} L_{\mathcal{f}}(W_0) $ on the first task, where $ L_{\mathcal{f}} $ is the full batch gradient over the first dataset, and then perform Singular Value Decomposition (SVD) to obtain $ U, S, V \leftarrow \nabla_{W_0} L_{\mathcal{f}}(W_0)$. The initialization sets $ A_1 \gets V_{[1:r_1]} $ and the remaining basis as $ \mathcal{E}_1 \gets V_{[r+1:]} $.

The results in Table \ref{tab:different basis} show that although both random and LoRA-GA initialization methods ensure orthogonality, they perform worse than the standard orthogonal basis. In the case of random initialization, the basis vectors are orthogonal but misaligned with the input data, which limits the model’s ability to leverage useful features and hinders learning performance.

In LoRA-GA, while the SVD operation effectively captures the dominant patterns from the first task and accelerates convergence, our experiments reveal a crucial drawback. Specifically, the LoRA selection mechanism tends to favor later singular vectors in the SVD decomposition. These later singular vectors are more likely to be nearly orthogonal to the input space relevant for future tasks. As a result, the new task’s LoRA components, which are selected from these vectors, become poorly aligned with the input features required for subsequent tasks. This misalignment leads to significant difficulties in adapting to new tasks, reflecting the stability–plasticity dilemma: while the SVD-based initialization ensures high stability for the first task, it compromises the plasticity needed for learning future tasks, ultimately degrading performance on later tasks.


In conclusion, while all three approaches enforce orthogonality, the standard orthogonal basis proves most effective, likely due to its natural alignment with the input space—potentially influenced by pretraining mechanisms such as dropout regularization~\cite{srivastava2014dropout}. Though not universally optimal, it effectively captures the intrinsic data structure, enabling more robust continual learning.

\begin{table}[h]
\centering
\begin{tabular}{ccccc}
\toprule
\textbf{$p$} & 1 & 2  & $\infty$ \\
\midrule
 Acc (\%) & 80.32 & 87.54  & 79.94 \\
\bottomrule
\end{tabular}
\caption{Comparison of different $p$ values on \textit{CIFAR-100}.}
\label{tab:different p}
\end{table}

\paragraph{Ablation Study on $p$.} We also design an ablation study on $p$ and $q$. By default, we set $p = 2$ to balance the contribution of potential weights in the future. To explore extreme cases, we set $p \rightarrow \infty$ and $p = 1$. When $p \rightarrow \infty$, Eq.~\eqref{eq:solveepsilon} becomes:
\begin{equation}\label{qeqone}
    \hat{\epsilon}_t(W_t) = \rho\ \text{sign}(\textbf{g}),
\end{equation}
which eliminates the magnitude of the gradient and retains only its direction. When $p=1$, Eq. \eqref{eq:solveepsilon} transforms to 
\begin{equation}
    {\hat{\epsilon}(W_{ij})} = 
\begin{cases} 
\frac{1}{\textbf{g}_{ij}}, & \text{if}\ \textbf{g}_{ij} = \max(\textbf{g}) \\
0, & \text{otherwise}
\label{peqone}
\end{cases}
\end{equation}

The results, shown in Table~\ref{tab:different p}, demonstrate that both extreme configurations lead to unstable outcomes. From a numerical perspective, the magnitude of Eq. \eqref{eq:solveepsilon} is much smaller than in Eq. \eqref{qeqone}, resulting in ${W}\ll\hat{\epsilon}(W)$ during the normal training of deep neural networks. This reasoning is similarly applicable to Eq. ~\eqref{peqone}, where $g_{ij} \ll 1$ implies that $W$ and ${W}_{ij} \ll \hat{\epsilon}(W_{ij})M_{ij}$, where $i$ and $j$ refer to the row and column indices corresponding to the maximum absolute value of $g$. Since our PLAN method involves a min-max problem, an imbalance in difficulty between the minimization and maximization components can lead to training instability. When gradient magnitudes vary significantly, the optimization may fail to converge or even diverge. This issue, which is common in min-max optimization \cite{DBLP:conf/iclr/ArjovskyB17}, arises when one component’s perturbations are disproportionately large or small, disrupting stable training.

\begin{table*}[t]
\centering
\setlength{\tabcolsep}{15pt}
\resizebox{0.85\linewidth}{!}{
\begin{tabular}{l|cc|cc}
\toprule
\multirow{2}{*}{\textbf{Method}} & \multicolumn{2}{c|}{\textit{CIFAR-100}} & \multicolumn{2}{c}{\textit{ImageNet-20}} \\
\cline{2-5}
& \raisebox{-0.25ex}[0pt][0pt]{Acc $\uparrow$} & \raisebox{-0.25ex}[0pt][0pt]{AAA $\uparrow$} & \raisebox{-0.25ex}[0pt][0pt]{Acc $\uparrow$} & \raisebox{-0.25ex}[0pt][0pt]{AAA $\uparrow$} \\
\midrule
L2P & 47.42 \footnotesize{(±1.12)} & 65.26 \footnotesize{(±0.59)} & 73.71 \footnotesize{(±0.27)} & 81.61 \footnotesize{(±0.43)} \\
Dual-Prompt & 57.38 \footnotesize{(±0.36)} & 65.26 \footnotesize{(±0.45)} & 69.61 \footnotesize{(±0.83)} & 76.92 \footnotesize{(±1.19)} \\
CODA-Prompt & 59.57 \footnotesize{(±0.33)} & 66.05 \footnotesize{(±0.30)} & \textbf{78.78} \footnotesize{(±0.65)} & \textbf{86.63} \footnotesize{(±0.43)} \\
\textit{Inc-LoRA} & 64.10 \footnotesize{(±0.45)} & 72.33 \footnotesize{(±0.29)} & 71.20 \footnotesize{(±0.80)} & 82.47 \footnotesize{(±0.26)} \\
O-LoRA & 53.26 \footnotesize{(±0.59)} & 63.03 \footnotesize{(±1.25)} & 72.75 \footnotesize{(±1.55)} & 81.46 \footnotesize{(±1.27)} \\
InfLoRA & 65.28 \footnotesize{(±0.37)} & 74.11 \footnotesize{(±0.47)} & 78.11 \footnotesize{(±0.35)} & 86.47 \footnotesize{(±0.17)} \\
\rowcolor{myrowcolor} PLAN (ours) & \textbf{65.93} \footnotesize{(±0.90)} & \textbf{74.46} \footnotesize{(±1.03)} & 78.39 \footnotesize{(±0.49)} & 86.61 \footnotesize{(±0.86)} \\
\bottomrule
\end{tabular}
}
\caption{Comparison of different methods on \textit{CIFAR-100} and \textit{ImageNet-R} ($N=20$) with iBOT-1k.}
\label{tab:ibot}
\end{table*}

\begin{table*}[t]
\centering
\setlength{\tabcolsep}{12pt}
\resizebox{0.85\linewidth}{!}{
\begin{tabular}{l|cc|cc}
\toprule
\multirow{2}{*}{\textbf{Method}} & \multicolumn{2}{c|}{\textit{CIFAR-100}} & \multicolumn{2}{c}{\textit{ImageNet-20}} \\
\cline{2-5}
& \raisebox{-0.25ex}[0pt][0pt]{Acc $\uparrow$} & \raisebox{-0.25ex}[0pt][0pt]{AAA $\uparrow$} & \raisebox{-0.25ex}[0pt][0pt]{Acc $\uparrow$} & \raisebox{-0.25ex}[0pt][0pt]{AAA $\uparrow$} \\
\midrule
Random Orthgonal Basis & 81.21 \footnotesize{(±0.45)}& 89.54 \footnotesize{(±0.35)}&69.42	\footnotesize{(±0.48)} & 77.94\footnotesize{(±0.87)}\\
LoRA-GA Basis& 84.30 \footnotesize{(±0.23)}& 91.14 \footnotesize{(±0.40)}& 69.40 \footnotesize{(±0.40)}& 77.51 \footnotesize{(±0.82)}\\
Standard Basis (ours) & \textbf{87.54} \footnotesize{(±0.31)} & \textbf{92.21} \footnotesize{(±0.35)} & \textbf{71.06} \footnotesize{(±0.42)} & \textbf{77.93} \footnotesize{(±0.56)} \\
\bottomrule
\end{tabular}
}
\caption{Comparison of different basis set $\mathcal{E}$ initialization methods.}
\vskip -0.05in
\label{tab:different basis}
\end{table*}

\begin{table}[h]
\centering
\begin{tabular}{|l|c|c|}
\hline
\rowcolor{gray!30} \textbf{Method} & \textbf{EP (M)} & \textbf{SF (M)} \\
\hline
L2P  & 1.85 & 0 \\
DualPrompt  & 14.65 & 0 \\
CODA-Prompt  & 5.57 & 0 \\
\textit{Inc-LoRA} & 1.41 & 0 \\
O-LoRA  & 1.41 & 13.36 \\
InfLoRA  & 0.70 & 67.35 \\
PLAN (ours) & 0.70 & 0 \\
\hline
\end{tabular}
\caption{Comparison of methods by Expended Parameters (EP) and Stored Features (SF) in MB on \textit{ImageNet-R} (N=20). }
\label{tab:comparsion of store}
\end{table}

\subsection{Analysis of Parameter and Storage Efficiency}

We compare the trainable parameters and storage requirements of various CL methods. The results are shown in Table~\ref{tab:comparsion of store}. For prompt-based methods such as L2P, DualPrompt, and CODA-Prompt, the learnable parameters are incorporated into the added prompts, and the corresponding keys must be stored. Notably, O-LoRA~\cite{wang-etal-2023-orthogonal} requires storing the previous LoRA $A$ block to compute the orthogonal loss, while InfLoRA~\cite{liang2024inflora} necessitates storing the gradient space. In contrast, our PLAN method only requires storing a negligible number of basis indices., which significantly reduces the storage requirements for both training and inference.

\subsection{Analysis of Pre-trained Model}

We conducted experiments using a ViT-B/16 model pre-trained with iBOT \cite{zhou2021image}. All experimental settings, except for the choice of the pre-trained model, remain consistent with those outlined in Section \ref{sec:setup}. Table~\ref{tab:ibot} presents the results of different methods on CIFAR-100 and ImageNet-R ($N=20$). Upon comparing these results with those presented in Table 1, we observe that all methods utilizing self-supervised pre-trained models yield lower performance compared to their counterparts with supervised pre-trained models. However, in this context, we find that PLAN either outperforms most methods or shows comparable performance, demonstrating its robustness even with self-supervised pre-training.

\section{Conclusion}

In this paper, we introduce PLAN (Proactive Low-Rank Allocation), a novel continual learning method that enhances LoRA with forward-looking subspace allocation and robust training objective.
Unlike existing approaches that passively enforce orthogonality to mitigate interference, PLAN anticipates future conflicts and proactively assigns task-specific subspaces, ensuring interference-free knowledge retention while simultaneously fostering adaptability through perturbation-aware optimization. By strategically preparing for future updates rather than simply reacting to past interference, PLAN provides a more effective solution to the stability–plasticity dilemma. 

\paragraph{Limitations and Future Work.} While PLAN shows strong performance, we identify several avenues for future research. First, its strict orthogonality, while excellent for preventing forgetting, does not explicitly promote positive backward transfer; exploring methods to selectively relax orthogonality could be beneficial. Second, our experiments primarily focused on ViT-based models; extending and evaluating PLAN on other architectures like ConvNets or different data modalities would be a valuable next step. Finally, while the standard basis proved effective, exploring adaptive basis generation for highly heterogeneous task sequences remains an interesting direction.

\section*{Acknowledgement}

This work was supported by National Key Research and Development Program of China (No. 2022ZD0160300) and NSFC grant (No. 62136005),

{
    \small
    \bibliographystyle{ieeenat_fullname}
    \bibliography{main}
}
\newpage
\appendix
\section{Appendix}
\label{sec:appendix}

\subsection{Analysis on Hyperparameter S}
\label{sec:appendix_analysis_s}

\begin{figure}[h]
\centering
\includegraphics[width=0.7\linewidth]{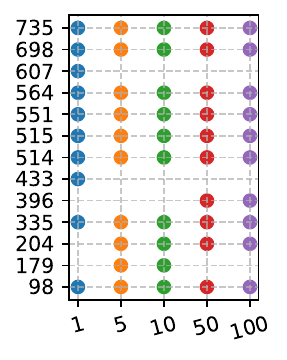}
\caption{Top-10 selected basis indices (y-axis) for the next task as a function of the sliding window size $S$ (x-axis). Each colored line tracks a specific basis index. The selection stabilizes with a small $S$.}
\label{fig:appendix_s_analysis}
\end{figure}

The parameter $S$ in Eq.~(12) denotes the size of the sliding window used to accumulate frequencies for basis selection. To analyze its impact, we computed the top-10 selected basis indices in the final layer for a task on CIFAR-100, varying $S$ from 1 to 100. As shown in Figure~\ref{fig:appendix_s_analysis}, the set of selected indices stabilizes very quickly. The indices chosen with a small window (e.g., $S=50$) are nearly identical to those chosen with a full window ($S=100$, equivalent to all training steps). This indicates that a short-term memory of perturbation sensitivity is sufficient for robust basis selection, justifying our choice of $S=50$ for efficiency.

\subsection{Analysis on Hyperparameter $\rho$}
\label{sec:appendix_analysis_rho}

The hyperparameter $\rho$ controls the perturbation magnitude in our min-max objective (Eq.~(8)). We performed an ablation study on \textit{ImageNet-R} ($N=5$) to determine its optimal value. The results are shown in Table~\ref{tab:appendix_rho}. A value of $\rho = 0.01$ provides the best balance, leading to the highest performance. Larger values (e.g., 0.1) or smaller values (e.g., 0.001) resulted in slightly degraded performance, demonstrating the model's sensitivity to this parameter.

\begin{table}[h]
\centering
\caption{Ablation study on $\rho$ on \textit{ImageNet-R} ($N=5$).}
\label{tab:appendix_rho}
\resizebox{0.7\linewidth}{!}{
\begin{tabular}{ccc}
\toprule
\textbf{$\rho$} & \textbf{Acc (\%)} & \textbf{AAA (\%)} \\
\midrule
0.1 & 75.23 & 78.94 \\
\textbf{0.01} & \textbf{77.79} & \textbf{81.93} \\
0.001 & 76.38 & 79.36 \\
\bottomrule
\end{tabular}
}
\end{table}

\subsection{Discussion on Backward Transfer}
Our work prioritizes stability, using strict orthogonal subspaces to effectively mitigate catastrophic forgetting, a success confirmed by our strong empirical results. This focus on interference prevention, however, means that PLAN does not explicitly facilitate positive backward transfer. Given that significant backward transfer is rarely observed in rehearsal-free CL, this represents a deliberate design choice. For future work, we believe exploring methods to dynamically adjust the degree of orthogonality could unlock opportunities for knowledge sharing across tasks while maintaining robustness.

\end{document}